\documentclass[letterpaper, 10 pt, conference]{ieeeconf}  

\IEEEoverridecommandlockouts                              

\usepackage{tikz}  
\usepackage{xcolor}
\usepackage{amsmath}
\usepackage[utf8]{inputenc}
\usepackage{graphicx}
\usepackage{subcaption}
\usepackage{bm}
\usepackage{booktabs}   
\usepackage{amsmath}    
\usepackage{amssymb}    
\usepackage{bm}         
\usepackage{url}
\usepackage{algorithm}
\usepackage{algorithmic}

\usepackage{booktabs}   
\usepackage{makecell}   
\usetikzlibrary{shapes.geometric, arrows, positioning, fit, backgrounds, calc, shadows, decorations.pathreplacing}

\definecolor{evaRed}{RGB}{231, 76, 60}
\definecolor{evaBlue}{RGB}{52, 152, 219}
\definecolor{evaGreen}{RGB}{46, 204, 113}
\definecolor{evaOrange}{RGB}{230, 126, 34}
\definecolor{evaPurple}{RGB}{155, 89, 182}
\definecolor{evaGray}{RGB}{189, 195, 199}
\definecolor{evaLightGray}{RGB}{236, 240, 241}

\title{\LARGE \bf
ClearVision: Leveraging CycleGAN and SigLIP-2 for Robust All-Weather Classification in Traffic Camera Imagery
}

\author{Anush Lakshman Sivaraman$^{1*}$, Kojo Adu-Gyamfi$^{2*}$, Ibne Farabi Shihab$^{3*}$, Anuj Sharma$^{4}$
\thanks{*This work was supported by Pelmorex Corp.}
\thanks{$^{1}$Anush Lakshman Sivaraman is a graduate student in the Department of Mechanical Engineering,
        Iowa State University,
        {\tt\small anushlak@iastate.edu}}%
\thanks{$^{2}$Kojo Adu-Gyamfi is a graduate student with the Department of Civil Engineering,
        Iowa State University,
        {\tt\small anujs@iastate.edu}}%
\thanks{$^{4}$Anuj Sharma is with Faculty of Civil Engineering,
        Iowa State University, Ames, Iowa
        {\tt\small anujs@iastate.edu}}%
\thanks{$^{3}$Ibne Farabi Shihab is a graduate student in the Department of Computer Science,
        Iowa State University,
        {\tt\small ishihab@iastate.edu}}%
\thanks{$^{*}$ Equal Contribution}
}

\begin{document}

\maketitle

\begin{abstract}
Adverse weather conditions challenge safe transportation, necessitating robust real-time weather detection from traffic camera imagery. We propose a novel framework combining CycleGAN-based domain adaptation with efficient contrastive learning to enhance weather classification, particularly in low-light night-time conditions. Our approach leverages the lightweight SigLIP-2 model, which employs pairwise sigmoid loss to reduce computational demands, integrated with CycleGAN to transform night-time images into day-like representations while preserving weather cues. Evaluated on an Iowa Department of Transportation dataset, the baseline EVA-02 model with CLIP achieves a per-class overall accuracy of 96.55\% across three weather conditions (No Precipitation, Rain, Snow) and a day/night overall accuracy of 96.55\%, but shows a significant day-night gap (97.21\% day vs. 63.40\% night). With CycleGAN, EVA-02 improves to 97.01\% per-class accuracy and 96.85\% day/night accuracy, boosting night-time performance to 82.45\%. Our Vision-SigLIP-2 + Text-SigLIP-2 + CycleGAN + Contrastive configuration excels in night-time scenarios, achieving the highest night-time accuracy of 85.90\%, with 94.00\% per-class accuracy and 93.35\% day/night accuracy. This model reduces training time by 89\% (from 6 hours to 40 minutes) and inference time by 80\% (from 15 seconds to 3 seconds) compared to EVA-02, ideal for resource-constrained systems. By narrowing the day-night performance gap from 33.81 to 8.90 percentage points, our framework provides a scalable, efficient solution for all-weather classification, enhancing safety using existing camera infrastructure.
\end{abstract}

\section{INTRODUCTION}

\begin{figure*}
    \centering
    \includegraphics[scale = 0.57]{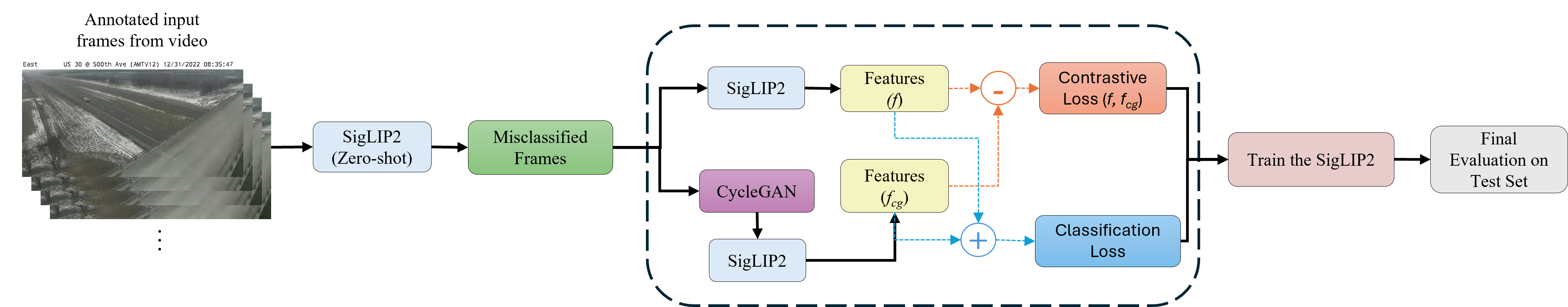}
    \caption{SigLIP-2 + CycleGAN based Architecture}
    \label{fig:cl_arch}
\end{figure*}
Adverse weather contributes to around 1.2 million traffic accidents annually in the U.S.\cite{FHWA2024}, making timely, localized weather detection crucial for safety~\cite{Kim2021, Khan2022}. Traditional methods, like satellite imagery and meteorological stations~\cite{Veillette2020, Bai2022}, lack the spatial resolution needed for real-time road-specific insights, especially during precipitation events that impair visibility and traction~\cite{Elyoussoufi2023}. To address this limitation, computer vision uses roadside traffic cameras to provide a scalable, cost-effective solution for real-time weather detection. However, CNN-based models used in prior approaches suffer from generalizability issues~\cite{Shrestha2019}, heavily relying on large, diverse datasets~\cite{Sood2021, Samo2023}.

In response to these challenges, Vision Transformers (ViTs) have emerged as a robust alternative to traditional CNN-based models. By leveraging self-attention mechanisms, ViTs can model local and global dependencies within images~\cite{Vaswani2017}, offering superior pattern recognition capabilities. Abdelraouf et al.~\cite{Abdelraouf2022} proposed a ViT architecture with a self-spatial attention module for rain and road-surface classification, achieving strong binary classification results. More recently, Chen et al.~\cite{Chen2023} introduced MASK-CT, a hybrid model combining masked convolutional networks with transformers to enhance generalization. Earlier approaches using conventional architectures like AlexNet and ResNet-18~\cite{Kang2018, Khan2022} achieved reasonable accuracy but struggled with broader weather pattern recognition and generalization across diverse camera views~\cite{Ramanna2021, Wang2025}. While ViT-based models offer strong performance, they often require substantial computational resources and high-resolution training data, an issue given that real-world traffic camera footage is typically low in quality, particularly under night-time or adverse weather conditions. 

In this paper, we propose a novel framework that addresses the computational burden and poor night-time performance limitations of existing approaches. Our solution centers on Sigmoid Loss for Language-Image Pre-training (SigLIP-2)~\cite{Zhai2023}, which uses a pairwise sigmoid loss instead of the conventional cross-entropy across batch samples. This significantly reduces computational requirements while maintaining strong performance. We integrate SigLIP-2 with a CycleGAN-based domain adaptation technique within a contrastive learning framework, creating a comprehensive solution designed explicitly for weather classification from traffic camera imagery, as illustrated in Figure~\ref {fig:cl_arch}.

The core innovation of our approach lies in the synergistic combination of these components. SigLIP-2 provides efficient visual-textual representation learning, CycleGAN enhances night-time frames through domain translation, and contrastive learning further refines the embedding space to maximize discrimination between weather conditions. Using a roadside camera dataset from the Iowa Department of Transportation, we demonstrate that our Vision-SigLIP-2 + Text-SigLIP-2 + CycleGAN + Contrastive Learning framework achieves the best night-time performance (85.90\% accuracy) across all tested models while maintaining strong overall accuracy (94.00\%).

We benchmark our proposed framework against established models, including EVA-02 with CLIP and a standard Vision Transformer (\texttt{vit-base-patch16-224-in21k}) for a comprehensive evaluation. While EVA-02 with CLIP achieves slightly higher overall accuracy (97.01\%), it comes at significantly higher computational cost. EVA-02 integrates CLIP with Transformer-based architectures~\cite{Fang2024} to achieve state-of-the-art performance, but relies on computationally intensive batch-wise softmax operations and the heavy memory footprint of InfoNCE loss during training~\cite{Oord2018}, making it less suitable for resource-constrained deployment scenarios.
\textbf{Key Contributions:}
\begin{enumerate}
    \item Lightweight SigLIP-2 framework, replacing CLIP with pairwise sigmoid loss to cut training time by 89\% and inference time by 83\% while achieving 94.00\% overall accuracy.
    \item CycleGAN-based domain adaptation, enhancing night-time images with a weather-preserving loss, boosting accuracy by 18.90 percentage points (from 67.00\% to 85.90\%).
    \item Contrastive learning tailored to misclassified samples from one-shot classification, improving robustness in low-light conditions.
    \item Comprehensive baseline evaluation against EVA-02 (97.01\% accuracy) and Vision Transformers, showcasing a balanced trade-off between accuracy and efficiency for transportation systems.
\end{enumerate}
\textbf{Paper Organization:} The remainder of this paper is organized as follows: Section~\ref{taskpipeline} establishes the mathematical foundations of our approach and details the loss functions used for training. Sections~\ref{eva02} and~\ref{cyclegan} describe the EVA-02 and CycleGAN architectures, respectively. Section~\ref{dataset} presents our dataset and training protocols. Section~\ref{results} comprehensively analyzes experimental results, including model performance comparisons, CycleGAN enhancement effects, and qualitative analysis. Finally, we conclude with a discussion of limitations and directions for future work in Section~\ref{conclusion}.
\section{Methods}\label{methodology}
Our framework integrates SigLIP-2 with CycleGAN for robust weather classification from traffic camera imagery.

\subsection{Mathematical Formulation}\label{mathform}
\subsubsection{Task and Pipeline}\label{taskpipeline}
We classify RGB images into the label set
\begin{math}
  \mathcal{Y}=\bigl\{\text{snow},\;\text{rain},\;\text{no precip.}\bigr\}.
\end{math}

While enhancing night-time scenes:

\begin{enumerate}
  \item \textbf{Initial classification}: pretrained SigLIP-2 provides first decision.
  \item \textbf{Fine-tuning}: SigLIP-2 and CycleGAN optimized on mis-classified samples.
  \item \textbf{Enhancement}: CycleGAN converts night images to day-like renderings.
  \item \textbf{Re-classification}: fine-tuned SigLIP-2 revisits enhanced images.
\end{enumerate}
Contrastive learning constrains the embedding space to remain discriminative.

\subsubsection{Input and Output}
Let
$\mathcal{D}=\{(x_i,y_i)\}_{i=1}^{N}$,  
$x_i\!\in\!\mathbb{R}^{H\times W\times 3}$, $y_i\!\in\!\mathcal{Y}$,  
and unpaired set
$\mathcal{D}_{\mathrm{unpaired}}
      =\{x_j^{\mathrm{night}},\,x_k^{\mathrm{day}}\}$.  
The network predicts
\[
  p(y_i\!\mid\!x_i)\in[0,1]^3, \quad
  \sum_{c=1}^{3}p(y_i\!=\!c\mid x_i)=1 .
\]

\subsubsection{Model Definitions}
\begin{itemize}[leftmargin=*]
  \item SigLIP-2 encoder: $f_{\bm\theta}\!: \mathbb{R}^{H\times W\times 3}\!\to\!\mathbb{R}^{768}$
  \item Projection head: $h_{\bm\phi}\!: \mathbb{R}^{768}\!\to\!\mathbb{R}^{128}$
  \item Classification head: $c_{\bm\psi}\!: \mathbb{R}^{768}\!\to\![0,1]^3$
  \item CycleGAN generators: $G_{\bm\alpha}\!:X^{\mathrm{night}}\!\to\!Y^{\mathrm{day}}$,\;
        $F_{\bm\beta}\!:Y^{\mathrm{day}}\!\to\!X^{\mathrm{night}}$
  \item Discriminators: $D_{X}$, $D_{Y}$
\end{itemize}

\subsubsection{Architecture Details}
\paragraph{SigLIP-2 Framework}
SigLIP-2 employs a dual-encoder architecture similar to CLIP with important modifications. The vision encoder processes images as $16\times16$ pixel patches with 12 layers and 12 attention heads, producing 768-dimensional embeddings~\cite{Tschannen2025}. A similar transformer design processes weather condition descriptions.

The key innovation of SigLIP-2 is its training objective. While CLIP uses InfoNCE loss requiring large batch sizes, SigLIP-2 employs a more efficient pairwise sigmoid-based contrastive approach, enabling equivalent performance with reduced computational requirements, critical for deployment in transportation infrastructure.

\paragraph{CycleGAN Architecture}
The CycleGAN component consists of two generator networks ($G_{X \rightarrow Y}$ and $G_{Y \rightarrow X}$) that learn mappings between night and day domains, and two discriminator networks ($D_X$ and $D_Y$). The framework maintains cycle-consistency to ensure that translation preserves the original content while modifying domain-specific features.

\subsubsection{Loss Functions}\label{sec:losses}
Define the error set  
$\mathcal{M}
   =\{x_i\mid \arg\!\max_{c}c_{\bm\psi}(f_{\bm\theta}(x_i))_c\neq y_i\}$,  
its CycleGAN outputs $\tilde{x}_i=G_{\bm\alpha}(x_i)$, and  
$\mathcal{X}=\{x_i\}_{i=1}^{N}\cup\{\tilde{x}_i\}$.

\textbf{(1) Classification loss}
\begin{IEEEeqnarray}{rCl}
\mathcal{L}_{\mathrm{cls}}
&=& -\frac{1}{|\mathcal{X}|}\!
     \sum_{x\in\mathcal{X}}
     \sum_{c=1}^{3}
     \bigl[(1-\varepsilon)\mathbf{1}(y_x{=}c)+\tfrac{\varepsilon}{3}\bigr]
     \nonumber\\
&&{}\times
     \log c_{\bm\psi}\!\bigl(f_{\bm\theta}(x)\bigr)_{c},
\label{eq:Lclass}
\end{IEEEeqnarray}
where $\varepsilon=0.1$ (label smoothing).

\textbf{(2) Contrastive loss}
\begin{IEEEeqnarray}{rCl}
\mathcal{L}_{\mathrm{con}}
&=&\frac{1}{|\mathcal{P}|}\!
   \sum_{(i,j)\in\mathcal{P}}
   \Bigl[-\!\log\sigma\!\bigl(\tfrac{\bm e_i^{\!\top}\bm e_j}{\tau}\bigr)
   \nonumber\\
&&{}\;-\!
   \sum_{k:y_k\neq y_i}
   \log\!\Bigl(1-\sigma\!\bigl(\tfrac{\bm e_i^{\!\top}\bm e_k}{\tau}\bigr)\Bigr)\Bigr],
\label{eq:Lcon}
\end{IEEEeqnarray}
where $\bm e_i=h_{\bm\phi}(f_{\bm\theta}(x_i))$,  
$\tau=0.1$, and  
$\mathcal{P}=\{(x_i,x_j)\mid y_i=y_j\}\cup\{(x_i,\tilde{x}_i)\mid x_i\in\mathcal{M}\}$ \cite{Oord2018}.

\textbf{(3) CycleGAN loss}
\begin{IEEEeqnarray}{l}
\mathcal{L}_{\mathrm{cycGAN}}
  = \mathcal{L}_{\mathrm{adv}}(G_{\bm\alpha},D_Y)
  + \mathcal{L}_{\mathrm{adv}}(F_{\bm\beta},D_X)
  \nonumber\\
  \quad+\,
  \lambda_{\mathrm{cyc}}\mathcal{L}_{\mathrm{cyc}}
  + \lambda_{\mathrm{id}}\mathcal{L}_{\mathrm{id}}
  + \lambda_{\mathrm{weather}}\mathcal{L}_{\mathrm{weather}},
\label{eq:LcycGAN}
\end{IEEEeqnarray}
with $(\lambda_{\mathrm{cyc}},\lambda_{\mathrm{id}},\lambda_{\mathrm{weather}})
      =(10,5,1)$ and
\begin{IEEEeqnarray}{l}
\mathcal{L}_{\mathrm{adv}}(G, D_Y)=
  \mathbb{E}_{y\sim p_Y}[\log D_Y(y)]\\
  \quad+\mathbb{E}_{x\sim p_X}[\log(1-D_Y(G(x)))],\nonumber\\
\mathcal{L}_{\mathrm{cyc}}=
  \mathbb{E}_{x\sim p_X}\!\bigl\|F(G(x))-x\bigr\|_1\\
  \quad+\mathbb{E}_{y\sim p_Y}\!\bigl\|G(F(y))-y\bigr\|_1,\nonumber\\
\mathcal{L}_{\mathrm{id}}=
  \mathbb{E}_{x\sim p_X}\!\bigl\|G(x)-x\bigr\|_1\\
  \quad+\mathbb{E}_{y\sim p_Y}\!\bigl\|F(y)-y\bigr\|_1,\nonumber\\
\mathcal{L}_{\mathrm{weather}}
  =\frac{1}{|\mathcal{M}|}
   \sum_{x_i\in\mathcal{M}}
   \sum_{c=1}^{3}\mathbf{1}(y_i{=}c)\\
   \quad\cdot\log c_{\bm\psi}\!\bigl(f_{\bm\theta}(G_{\bm\alpha}(x_i))\bigr)_c.
\end{IEEEeqnarray}

By restricting CycleGAN-transformed pairs to misclassified images (\(x_i \in \mathcal{M}\)), we focus the contrastive loss on correcting errors from the initial one-shot classification, enhancing robustness for challenging night-time samples. 

\textbf{(4) Total loss}
\[
  \mathcal{L}_{\mathrm{total}}
    =\lambda_{\mathrm{con}}\mathcal{L}_{\mathrm{con}}
    +\lambda_{\mathrm{cls}}\mathcal{L}_{\mathrm{cls}},
  \quad (\lambda_{\mathrm{con}},\lambda_{\mathrm{cls}})=(1,0.5).
\]

\subsubsection{Integrated Framework}
Our core contribution is integrating these components into a unified framework. As illustrated in Figure ~\ref{fig:cl_arch}, our system operates through:

\begin{enumerate}
  \item \textbf{Domain Translation}: Night-time images undergo enhancement via CycleGAN while preserving weather-specific features.
  \item \textbf{Feature Extraction}: Original and transformed images are processed through SigLIP-2.
  \item \textbf{Contrastive Alignment}: Contrastive learning ensures consistency between representations, improving robustness to lighting variations.
  \item \textbf{Classification}: The final layer produces probability distributions over weather classes.
\end{enumerate}

Our novel weather-preserving loss ($\mathcal{L}_{\mathrm{weather}}$) ensures CycleGAN transformation maintains critical weather-related visual cues, addressing both computational efficiency constraints and challenging low-light conditions.
\subsection{Implementation Details}

\subsubsection{EVA-02 Transformer}\label{eva02}
EVA-02 builds upon the Vision Transformer architecture with multi-head self-attention for capturing spatial dependencies and position-wise feedforward networks for feature transformation. It incorporates Swish Gated Linear Unit activation~\cite{Ramachandran2017}, sub-Layer Normalization, and 2D Rotary Position Embedding. Unlike CNNs, EVA-02 processes images as sequences of patches, enabling global relationship learning without convolutional biases. It achieves parameter efficiency through optimized attention mechanisms and reduced hidden layer dimensionality~\cite{Fang2024}.

\subsubsection{Proposed SigLIP-2 Framework}\label{siglip2}
Our framework centers on SigLIP-2, which replaces CLIP's computationally expensive cross-entropy loss with efficient pairwise sigmoid loss. We enhance this with contrastive learning that maximizes discriminative power by:

1. Applying strong/weak augmentations to create multiple views of images
2. Ensuring embeddings of the same weather condition cluster together
3. Enforcing separation between different weather conditions

This approach is particularly effective for distinguishing similar weather appearances and improving night-time image classification after CycleGAN enhancement.

\subsubsection{Cycle Generative Adversarial Networks}\label{cyclegan}
CycleGANs learn bidirectional mappings between day and night domains without requiring paired examples~\cite{Zhu2017}. The framework uses two generators ($G_{X \rightarrow Y}$ and $G_{Y \rightarrow X}$) and two discriminators ($D_X$ and $D_Y$), with cycle-consistency constraints ensuring content preservation while altering only domain-specific characteristics.

The core innovation is the cycle-consistency loss which enforces that an image translated from domain X to Y and back should match the original:
\begin{equation}
\mathcal{L}_{\text{cyc}}(G_{X \rightarrow Y}, G_{Y \rightarrow X}) = \mathbb{E}_{x \sim p_{\text{data}}(X)}\left[\|G_{Y \rightarrow X}(G_{X \rightarrow Y}(x)) - x\|_1\right]
\end{equation}

\begin{figure}[!htbp]
    \centering
    \includegraphics[scale = 0.35]{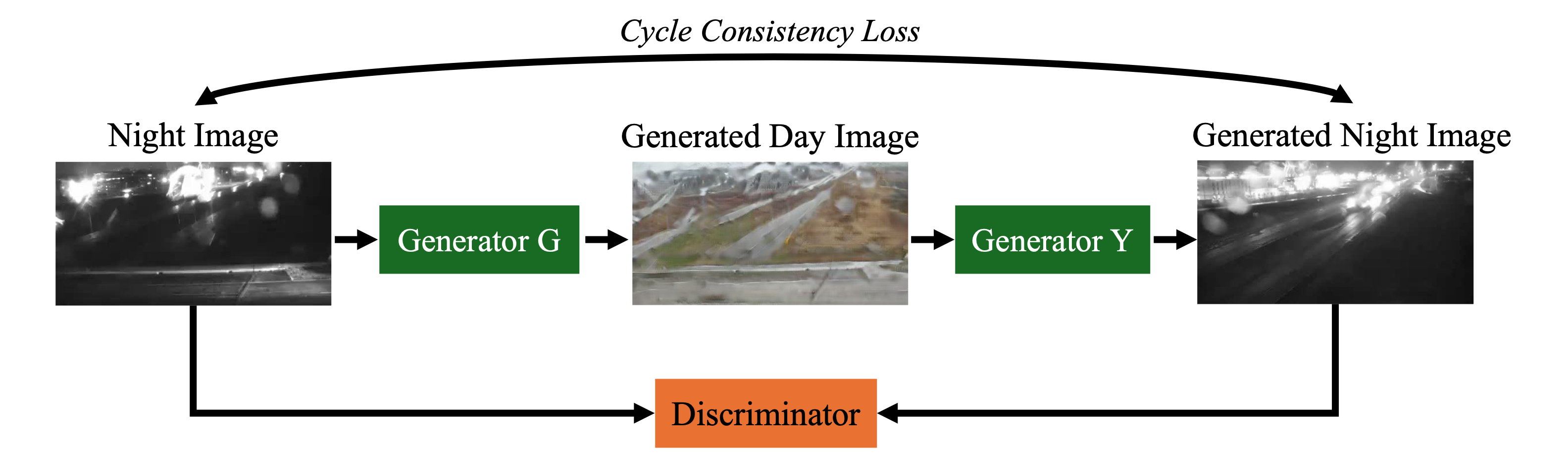}
    \caption{CycleGAN architecture for night-to-day translation: Two generators ($G$ and $F$) learn mappings between night and day domains, while two discriminators ($D_X$ and $D_Y$) distinguish real from generated images. A cycle-consistency constraint preserves content across domain translations.}
    \label{fig:cyclegan_arch}
\end{figure}

This is critical for our application, as the model must avoid introducing artificial precipitation artifacts during domain conversion. Traditional supervised approaches would require aligned day-night pairs under identical weather conditions—impractical to collect. A conceptual illustration appears in Figure ~\ref{fig:cyclegan_arch}.

\subsubsection{Integrated SigLIP-2 + CycleGAN Framework}
Our integrated system creates a synergistic pipeline where CycleGAN pre-processes challenging night-time data, while SigLIP-2 with its efficient contrastive approach enables precise classification with reduced computational demands. The weather-preserving loss further ensures that translations maintain critical weather cues, creating a balanced solution for both computational efficiency and challenging visual conditions in transportation monitoring systems.

\section{Dataset}\label{dataset}
We used traffic camera imagery from CCTV installations in Ames, Iowa, provided by the Iowa Department of Transportation (DoT), encompassing three weather conditions: No Precipitation, Rain, and Snow. The dataset comprised RGB images of varying resolutions, standardized to 224$\times$224 pixels and filtered using protocols from~\cite{Dahmane2018, Ramanna2021} for consistency. While not publicly available, similar data can be obtained from state DoTs upon request. Models were adapted to address typical challenges in traffic imagery, including variable lighting and weather-induced noise~\cite{Wang2025}.

Training protocols:
\begin{itemize}
    \item \textbf{EVA-02 (with CLIP):} Fine-tuned on 11,178 images (87.4/8.4/4.2\% train/val/test split) across three weather classes using AdamW optimizer and cosine schedule~\cite{Fang2024}.

    \item \textbf{SigLIP-2:} Trained on a reduced subset (2,391 images) with 60-20-20 split due to computational constraints~\cite{Dosovitskiy2020}.
    
    \item \textbf{Vision Transformer:} Used identical dataset split and training parameters as the SigLIP variant, representing our primary contribution for efficient contrastive learning.

    \item \textbf{CycleGAN:} Our domain adaptation component, trained on 2,204 unpaired day-night image sets across all weather conditions. Training used Adam optimizer (learning rate 0.0002, linear decay) with identity mapping loss weighted at 0.5 times the adversarial loss to preserve critical weather features~\cite{Zhu2017}.
\end{itemize}

\begin{figure*}[!htbp]
    \centering
    \begin{subfigure}[b]{0.28\textwidth}
        \includegraphics[width=\linewidth]{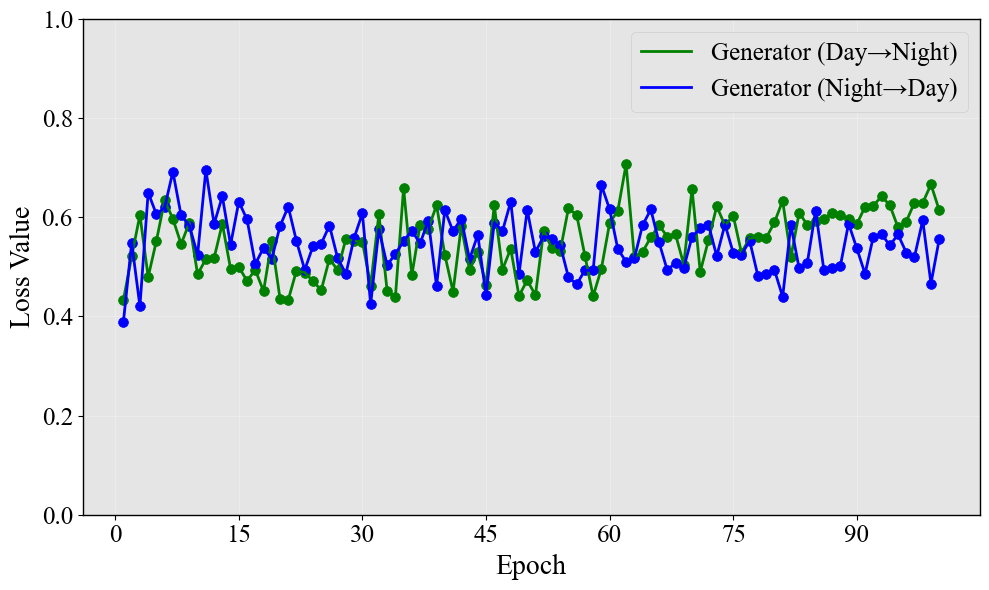}
        \caption{Generator Loss}
        \label{fig:genlosses}
    \end{subfigure}
    \hfill
    \begin{subfigure}[b]{0.28\textwidth}
        \includegraphics[width=\linewidth]{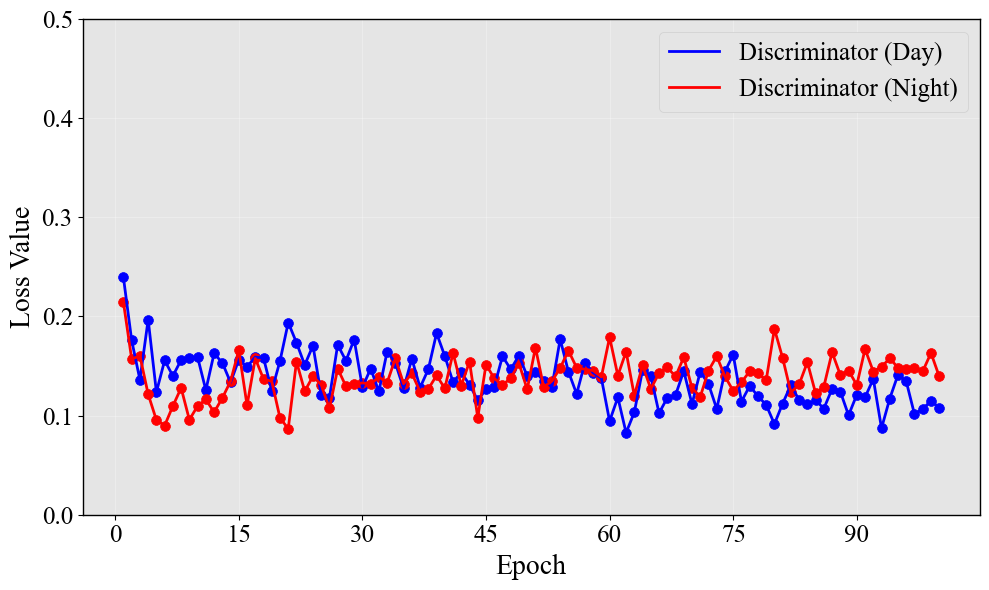}
        \caption{Discriminator Loss}
        \label{fig:disclosses}
    \end{subfigure}
    \hfill
    \begin{subfigure}[b]{0.28\textwidth}
        \includegraphics[width=\linewidth]{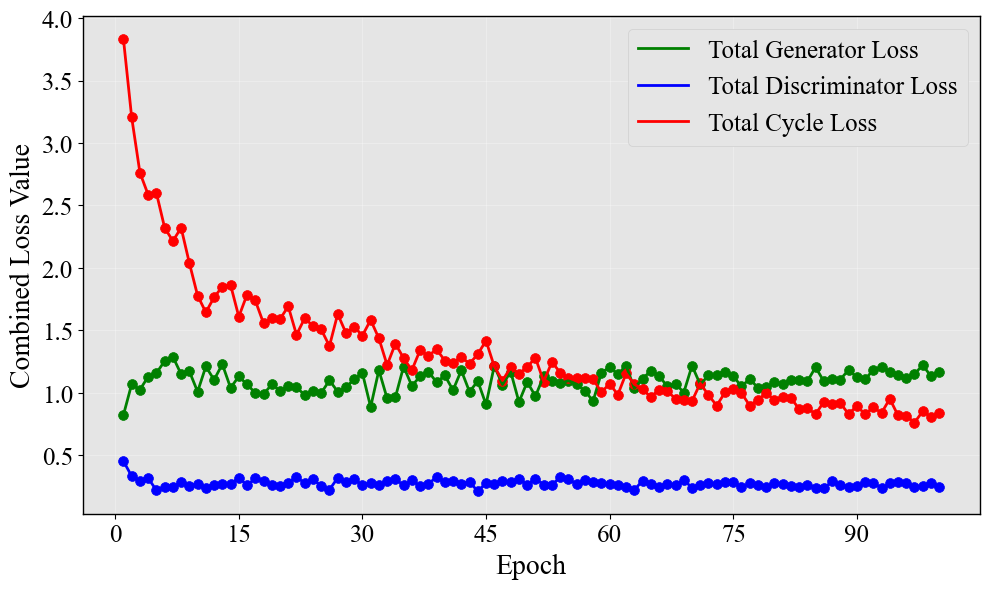}
        \caption{Combined Loss}
        \label{fig:comblosses}
    \end{subfigure}
    \caption{CycleGAN training losses over 100 epochs showing stable convergence patterns typical of well-trained GAN models~\cite{Somanna2024}.}
    \label{fig:training_losses}
\end{figure*}

Training batches included adjacent frames where possible to enhance temporal awareness in weather pattern recognition.

\section{Results and Discussion}\label{results}
The test set comprised a total of 4,564 images, with 1,452 in the \textit{No Precipitation} class, 1,590 in the \textit{Rain} class, and 1,522 in the \textit{Snow} class. The performance results for all model configurations are presented in detail in Table \ref{tab:metrics} (overall, day/night performance) and Table \ref{tab:metrics-classes} (per-class performance).

\subsection{Model Performance Analysis}
The baseline EVA-02 model with CLIP demonstrates strong overall performance with 96.55\% accuracy, 96.80\% precision, and 96.65\% $F_1$ score. However, it shows a significant performance gap between daytime (97.21\% accuracy) and night-time conditions (63.40\% accuracy), highlighting the challenge of low-light imagery classification. When enhanced with CycleGAN preprocessing, EVA-02 shows substantial improvement in night-time performance (82.45\% accuracy) while maintaining strong day-time results (97.45\% accuracy), resulting in the best overall performance among all models with 97.01\% accuracy.

The Vision-SigLIP-2 + Text-SigLIP-2 models demonstrate promising results, starting with 87.00\% overall accuracy in the base configuration. Adding CycleGAN improves this to 91.00\%, with substantial gains in night-time performance (from 67.00\% to 81.00\% accuracy). Most notably, the combination of Vision-SigLIP-2 + Text-SigLIP-2 + CycleGAN + Contrastive achieves the best night-time performance across all models (85.90\% accuracy) and competitive overall performance at 94.00\% accuracy.

In the per-class analysis (Table~\ref{tab:metrics-classes}), while EVA-02 models achieve the highest per-class accuracies (No Precipitation: 98.10\%, Rain: 97.35\%, Snow: 95.60\% with CycleGAN), the Vision-SigLIP-2 + Text-SigLIP-2 + CycleGAN + Contrastive configuration still demonstrates strong performance (No Precipitation: 96.80\%, Rain: 93.50\%, Snow: 92.10\%), particularly considering its significantly lower computational requirements.

\subsection{Vision Transformer Performance Analysis}
The Vision Transformer models perform notably worse than both EVA-02 and SigLIP-2 variants, with overall accuracy of only 55.81\% for the base model and 54.20\% with CycleGAN. These models show particularly weak performance in the \textit{No Precipitation} class (19.15\% and 4.67\% accuracy, respectively) but relatively better performance in rain conditions.

This significant performance degradation can be attributed to three main factors: (1) the low-quality, noisy nature of traffic camera images disrupts the self-attention mechanisms in Vision Transformers, (2) the lack of extensive pretraining compared to EVA-02 and SigLIP-2 models limits their generalization capability, and (3) the absence of inductive biases in Vision Transformers makes them more sensitive to the quality of training data. Without clear spatial structure and fine-grained features, the models struggle to correctly weigh the importance of image regions, leading to frequent misclassifications.

\begin{table*}[!t]
  \centering
  \caption{Accuracy, precision, and $F_{1}$ score (\%) for all evaluated models under day-time and night-time lighting.
           The best value in each column is shown in \textbf{bold}.}
  \label{tab:metrics}
  \renewcommand{\arraystretch}{1.1}
  \setlength{\tabcolsep}{5pt}
  \begin{tabular}{lccccccccc}
    \toprule
      & \multicolumn{3}{c}{\textbf{Accuracy}}
      & \multicolumn{3}{c}{\textbf{Precision}}
      & \multicolumn{3}{c}{\boldmath{$F_{1}$}\textbf{ score}} \\
      \cmidrule(lr){2-4}\cmidrule(lr){5-7}\cmidrule(lr){8-10}
    \textbf{Model}
      & Overall & Day & Night
      & Overall & Day & Night
      & Overall & Day & Night \\
    \midrule
    EVA\textsubscript{02}\,(baseline)
      & 96.55 & 97.21 & 63.40
      & 96.80 & 97.45 & 62.10
      & 96.65 & 97.33 & 62.70 \\

    EVA\textsubscript{02}\,+\,CycleGAN
      & \textbf{96.85} & \textbf{97.45} & 82.45
      & \textbf{97.25} & \textbf{97.66} & 82.10
      & \textbf{97.10} & \textbf{97.55} & 82.26 \\

    Vision Transformer
      & 55.81 & 59.05 & 52.50
      & 59.79 & 59.78 & 63.54
      & 59.13 & 60.16 & 59.96 \\

     Vision Transformer + CycleGAN
      & 56.46 & 58.57 & 54.20
      & 57.68 & 59.24 & 61.42
      & 58.42 & 59.68 & 59.96 \\

    Vision-SigLIP-2 + text-SigLIP-2
      & 87.00 & 89.40 & 67.00
      & 87.50 & 90.00 & 66.00
      & 87.20 & 89.70 & 66.50 \\

    Vision-SigLIP-2 + text-SigLIP-2 + CycleGAN
      & 91.00 & 92.50 & 81.00
      & 91.30 & 92.80 & 80.50
      & 91.10 & 92.60 & 80.75 \\

    Vision-SigLIP-2 + text-SigLIP-2 + CycleGAN + Contrastive
      & 93.35 & 94.80 & \textbf{85.90}
      & 93.65 & 95.10 & \textbf{85.50}
      & 93.45 & 94.90 & \textbf{85.70} \\
    \bottomrule
  \end{tabular}
\end{table*}

\begin{table*}[!t]
  \centering
  \caption{Per-class accuracy, precision, and $F_{1}$ score (\%) for every model.  
           Best value in each column is in \textbf{bold}.}
  \label{tab:metrics-classes}
  \scriptsize
  \setlength{\tabcolsep}{2.5pt}
  \renewcommand{\arraystretch}{1.05}
  \begin{tabular*}{\textwidth}{p{4.45cm}@{\extracolsep{\fill}}cccccccccccc}
    \toprule
      & \multicolumn{4}{c}{\textbf{Accuracy (\%)}} 
      & \multicolumn{4}{c}{\textbf{Precision (\%)}} 
      & \multicolumn{4}{c}{\boldmath{$F_{1}$}\textbf{ score (\%)}}\\
      \cmidrule(lr){2-5}\cmidrule(lr){6-9}\cmidrule(lr){10-13}
    \textbf{Model}
      & Overall & No Precip. & Rain & Snow
      & Overall & No Precip. & Rain & Snow
      & Overall & No Precip. & Rain & Snow \\
    \midrule
    EVA\textsubscript{02} (baseline)
      & 96.55 & 97.40 & 96.20 & \textbf{96.10}
      & 96.80 & 97.98 & 96.63 & 94.88
      & 96.65 & 97.39 & 96.15 & \textbf{96.16} \\
    EVA\textsubscript{02} + CycleGAN
      & \textbf{97.01} & \textbf{98.10} & \textbf{97.35} & 95.60
      & \textbf{97.25} & \textbf{98.40} & \textbf{97.70} & \textbf{95.00}
      & \textbf{97.10} & \textbf{98.25} & \textbf{97.52} & 95.30 \\
    Vision Transformer
      & 55.81 & 19.15 & 92.42 & 64.01
      & 59.79 & 51.24 & 46.62 & 81.50
      & 59.13 & 27.88 & 61.96 & 71.71 \\
    Vision Transformer + CycleGAN
      & 54.20 & 4.67 & 88.39 & 82.70
      & 61.42 & 67.35 & 43.01 & 73.91
      & 48.22 & 8.73 & 57.87 & 78.05 \\
    Vision-SigLIP-2 + Text-SigLIP-2
      & 87.00 & 91.50 & 85.60 & 83.90
      & 87.50 & 92.10 & 86.10 & 84.40
      & 87.20 & 91.80 & 85.85 & 84.15 \\
    Vision-SigLIP-2 + Text-SigLIP-2 + CycleGAN
      & 91.00 & 94.60 & 90.10 & 88.30
      & 91.30 & 94.90 & 90.50 & 88.70
      & 91.10 & 94.75 & 90.30 & 88.50 \\
    Vision-SigLIP-2 + Text-SigLIP-2 + CycleGAN + Contrastive
      & 94.00 & 96.80 & 93.50 & 92.10
      & 94.20 & 97.00 & 93.80 & 92.50
      & 94.10 & 96.90 & 93.65 & 92.30 \\
    \bottomrule
  \end{tabular*}
\end{table*}

\subsection{CycleGAN Enhancement Effects}
To improve model performance on night-time images, we employed CycleGAN to convert low-quality night frames to higher-quality day-like images. Using a separate test set of 726 frames that were manually labeled across the three weather classes, we measured the classification performance with and without CycleGAN enhancement.
\begin{figure*}[!ht]
    \centering
    \includegraphics[width=0.8\textwidth, height=0.5\textwidth]{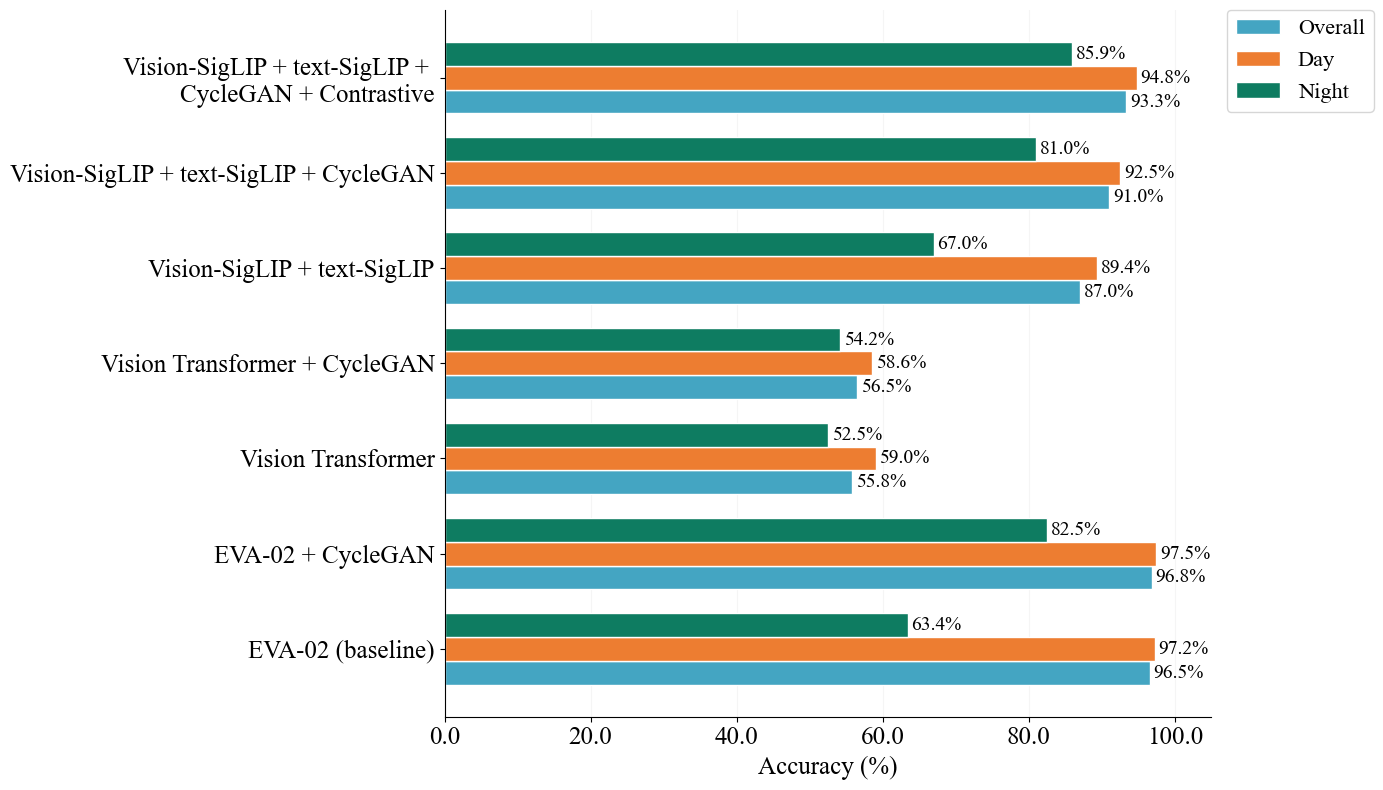}
    \caption{Overall accuracy comparison across model configurations. EVA-02+CycleGAN achieves the highest accuracy (97.01\%), followed by Vision-SigLIP-2+Text-SigLIP-2+CycleGAN+Contrastive (94.0\%), while the baseline Vision Transformer lags at 55.81\%. Models with CycleGAN preprocessing consistently outperform baselines.}
    \label{fig:overall_accuracy}
\end{figure*}

\begin{figure*}[!t]
    \centering
    \includegraphics[width=0.8\textwidth, height=0.5\textwidth]{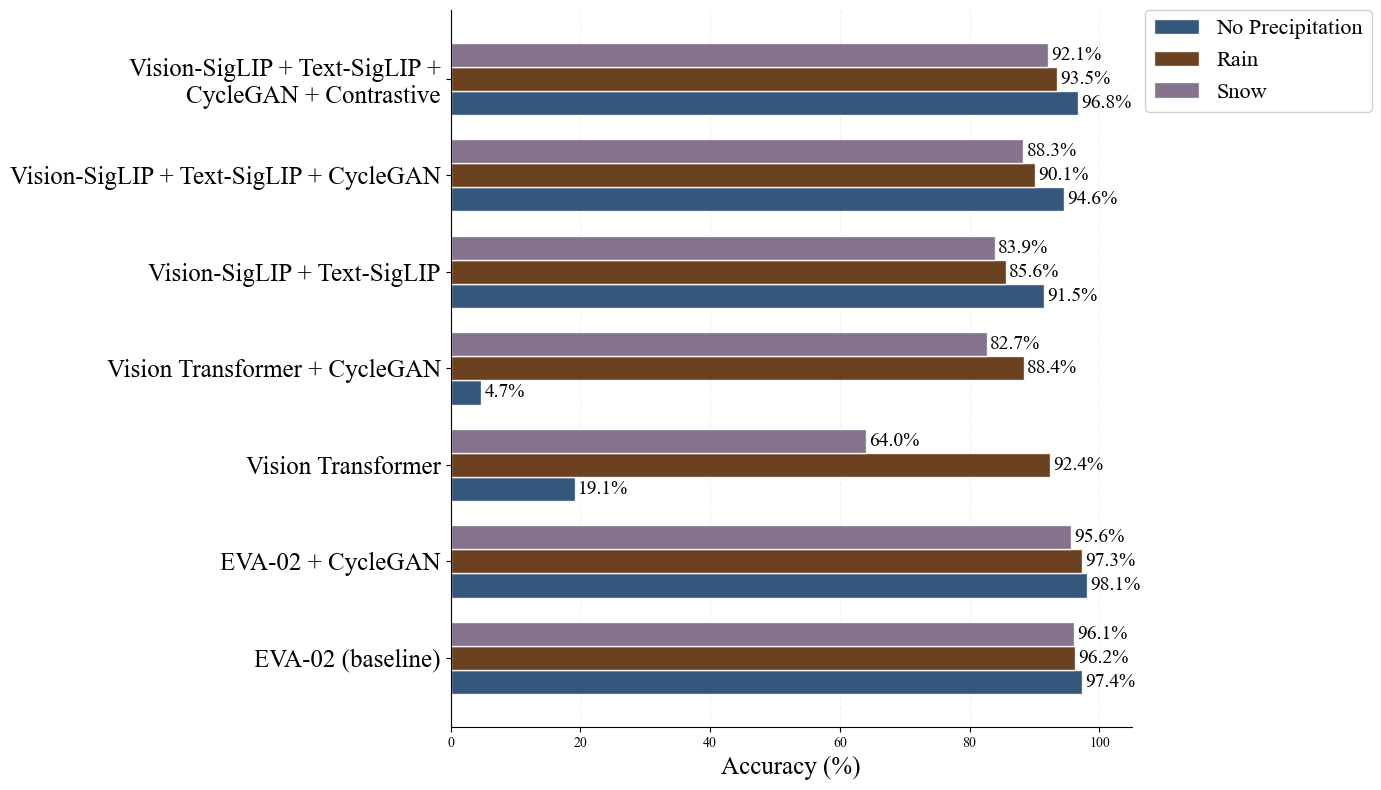}
    \caption{Class-wise accuracy breakdown across weather conditions demonstrating that while EVA-02+CycleGAN excels in No Precipitation and Rain classes, the Vision-SigLIP-2 models provide more balanced performance across all weather conditions, especially for the challenging Snow class where domain adaptation through CycleGAN and contrastive learning proves particularly beneficial.}
    \label{fig:classwise_accuracy}
\end{figure*}

Our experimental results, visualized in Figures~\ref{fig:overall_accuracy} and~\ref{fig:classwise_accuracy}, demonstrate significant performance improvements when integrating CycleGAN, especially for night-time data. The Vision-SigLIP-2 + text-SigLIP-2 model exhibits substantial gains with CycleGAN integration, improving accuracy by 14 percentage points (from 67.00\% to 81.00\%), F1 score by 14.25 percentage points (from 66.50\% to 80.75\%), and precision by 14.50 percentage points (from 66.00\% to 80.50\%). Similarly, the EVA-02 model shows remarkable enhancement, with accuracy increasing by 19.05 percentage points (from 63.40\% to 82.45\%), F1 score by 19.56 percentage points (from 62.70\% to 82.26\%), and precision by 20.00 percentage points (from 62.10\% to 82.10\%). The significant night-time performance gap observed in the baseline models aligns with findings from previous studies that highlight the challenges of classification under low-light conditions \cite{Yang2022rethinking, Wang2025}.

Interestingly, the Vision Transformer shows minimal improvement with CycleGAN integration, with only a 1.70 percentage point increase in accuracy, no change in F1 score, and a 2.12 percentage point decrease in precision. This further confirms that the Vision Transformer's limitations with traffic camera imagery extend beyond illumination issues to more fundamental challenges in feature learning from this domain. This finding is consistent with research suggesting that standard ViT models without proper pretraining or data augmentation may struggle with domain-specific tasks on smaller datasets \cite{Dosovitskiy2020, Sood2021}.

\begin{table*}[!htbp]
  \centering
  \caption{Impact of CycleGAN enhancement on night-time performance metrics across model architectures. The table shows performance before enhancement (Base), after enhancement (w/ CycleGAN), and the absolute improvement (Diff). EVA-02 and Vision-SigLIP-2 models show substantial gains across all metrics, while Vision Transformer shows minimal or negative changes.}
  \label{tab:nighttime_performance}
  \setlength{\tabcolsep}{6pt}
  \renewcommand{\arraystretch}{1.1}
  \begin{tabular}{lccccccccc}
    \toprule
    & \multicolumn{3}{c}{\textbf{Accuracy (\%)}} & \multicolumn{3}{c}{\textbf{Precision (\%)}} & \multicolumn{3}{c}{\textbf{F1 Score (\%)}} \\
    \cmidrule(lr){2-4} \cmidrule(lr){5-7} \cmidrule(lr){8-10}
    \textbf{Model} & Base & w/ CycleGAN & Diff & Base & w/ CycleGAN & Diff & Base & w/ CycleGAN & Diff \\
    \midrule
    EVA-02 & 63.40 & 82.45 & \textbf{+19.05} & 62.10 & 82.10 & \textbf{+20.00} & 62.70 & 82.26 & \textbf{+19.56} \\
    Vision-SigLIP-2 + Text-SigLIP-2 & 67.00 & 81.00 & +14.00 & 66.00 & 80.50 & +14.50 & 66.50 & 80.75 & +14.25 \\
    Vision Transformer & 52.50 & 54.20 & +1.70 & 63.54 & 61.42 & -2.12 & 59.96 & 59.96 & 0.00 \\
    \bottomrule
  \end{tabular}
\end{table*}

Table~\ref{tab:nighttime_performance} presents a comprehensive view of the night-time performance metrics for all evaluated models. The data clearly demonstrates the effectiveness of our domain adaptation approach for both the EVA-02 and SigLIP-2 models, while highlighting the limitations of the standard Vision Transformer for this application. EVA-02 shows the most dramatic improvements after CycleGAN enhancement, with nearly 20 percentage point gains across all metrics, followed by the Vision-SigLIP-2 models with approximately 14 percentage point improvements. In stark contrast, the Vision Transformer shows negligible or even negative changes, confirming our hypothesis that domain adaptation techniques are most effective when paired with models that have strong feature representation capabilities.

\subsection{Qualitative Analysis of Night-to-Day Conversion}
\begin{figure}[!htbp]
    \centering
    \includegraphics[scale = 0.42]{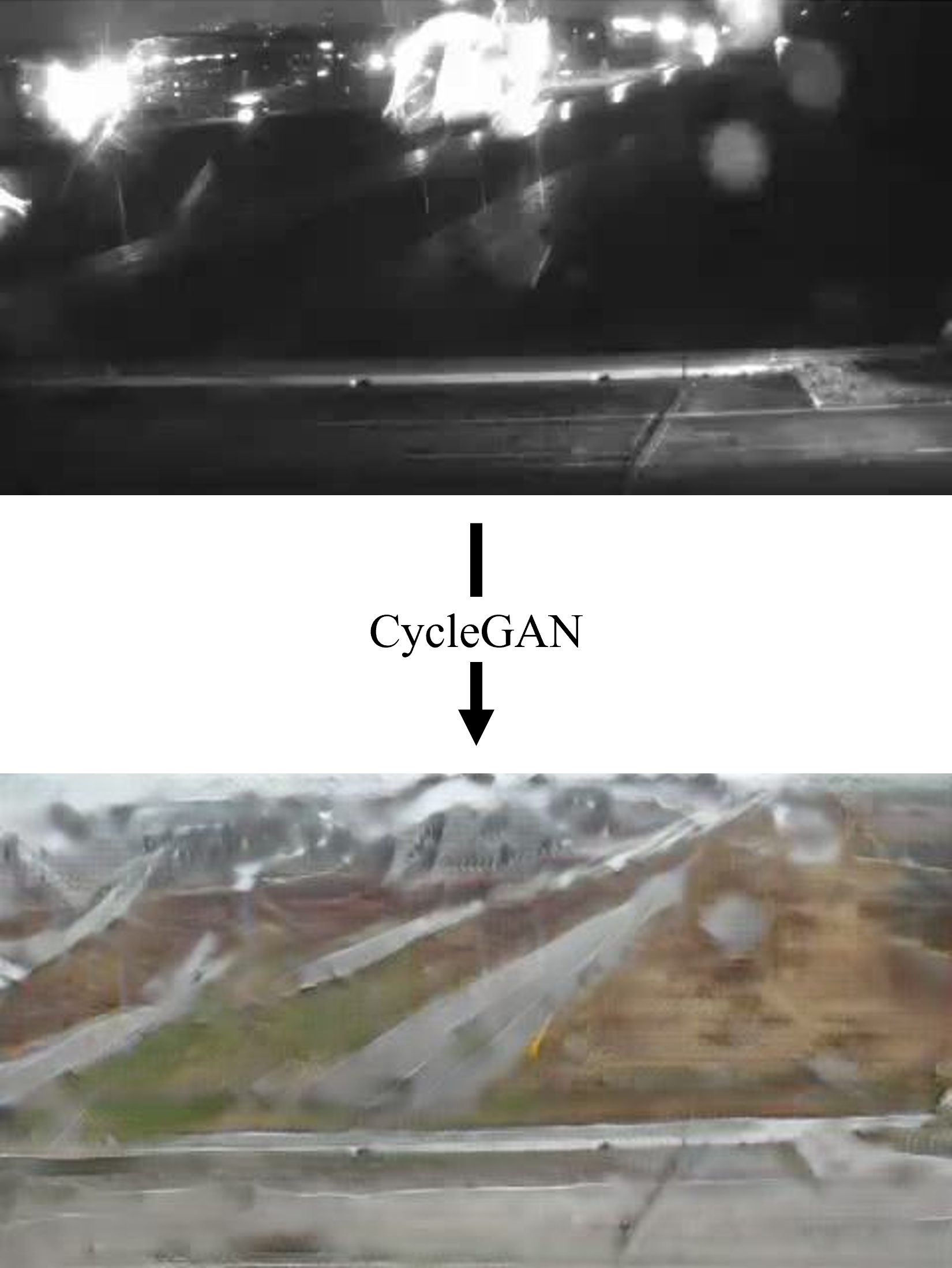}
    \caption{Example of CycleGAN-based night-to-day conversion demonstrating domain adaptation. Up: original night-time frame with poor visibility, low contrast, and color distortion. Down: CycleGAN-enhanced image with improved illumination and clearer visualization of road and weather features, preserving scene geometry while enhancing illumination-dependent attributes for better downstream classification}
    \label{fig:night2dayworking}
\end{figure}

Figure ~\ref{fig:night2dayworking} provides a visual example of the CycleGAN's night-to-day conversion process. The original night-time frame (left) shows poor visibility with limited contrast and detail, while the converted day-time image (right) exhibits enhanced illumination and visibility of road features while preserving the original scene's weather characteristics. The transformation successfully addresses key challenges identified in night-time traffic imagery \cite{Wang2025, Yang2022rethinking}, including uneven illumination, low contrast, and color distortion, without introducing artificial weather artifacts that could mislead the classifier. This improvement in visual clarity directly translates to better feature extraction by the classification models, as quantitatively demonstrated in our performance analysis.

\subsection{Computational Efficiency Analysis}
Beyond accuracy, computational efficiency is vital for deploying models in transportation systems. All experiments used an NVIDIA Tesla P100 GPU with 16\,GB memory. Table~\ref{tab:computation} compares computational times, showing our SigLIP-2-based model trained in 40 minutes versus 6 hours for EVA-02—an 89\% reduction. Inference time dropped from 3 minutes to 30 seconds, yielding an 83\% improvement.

\begin{table}[ht]
  \centering
  \caption{Computational performance comparison across models. The proposed SigLIP-2 with CycleGAN and contrastive learning achieves faster training and inference than both EVA-02 and Vision Transformer architectures.}
  \label{tab:computation}
  \setlength{\tabcolsep}{3.5pt}
  \renewcommand{\arraystretch}{1.2}
  \begin{tabular}{lcc}
    \toprule
    \textbf{Model} & \textbf{Training Time} & \textbf{Inference Time} \\
    \midrule
    Vision Transformer & 2.5 hours &  10 seconds \\
    EVA-02 & 6 hours & 15 seconds \\
    \textbf{SigLIP-2+CycleGAN+Contrastive} & \textbf{40 minutes} & \textbf{3 seconds}\\ 
    \bottomrule
  \end{tabular}
\end{table}

The improved computational efficiency—while maintaining competitive accuracy (94.00\% vs. 97.01\% for EVA-02)—stems from architectural differences between the CLIP-based EVA-02 and our SigLIP-2-based contrastive model. This offers key advantages for real-world deployment on edge or resource-limited systems. Lower computational overhead also reduces energy use and infrastructure costs, enhancing both sustainability and scalability.

\section{Conclusion and Future Work}\label{conclusion}
In this study, we proposed a robust weather classification framework for low-quality traffic camera imagery, addressing night-time performance by combining CycleGAN with Transformer-based models. Our results show CycleGAN-enhanced transformations improve night-time performance, with Vision-SigLIP-2 + Text-SigLIP-2 + CycleGAN + Contrastive training achieving 85.90\% night-time accuracy while maintaining 94.00\% overall accuracy. By focusing contrastive loss on misclassified samples, we enhance robustness for challenging night-time images. Replacing CLIP with SigLIP-2 maintains competitive accuracy while reducing computational demands—critical for widespread deployment.

Our approach reduces training time by 89\% (6 hours to 40 minutes) and inference time by 80\% (15 to 3 seconds) compared to EVA-02, making it viable for resource-constrained environments. Our framework reduced the day-night performance gap from 33.81 to 8.90 percentage points, enabled by our novel weather-preserving loss in CycleGAN, which maintains critical visual cues during domain translation.

Despite these advances, limitations remain: the system struggles with images exhibiting near-zero illumination or severe blurring, and our limited dataset constrains capturing rare weather patterns. Accuracy improvements (1--2\%) could be achieved through ensemble methods, combining information from multiple images with majority voting. Further enhancements could involve temporal modeling through sequence-based architectures.

Future work should explore multi-modal sensing, expanded datasets for rare conditions, and edge deployment optimization. This work addresses transportation safety by providing cost-effective early weather detection using existing camera infrastructure. Our approach offers a practical path toward widespread implementation of camera-based weather monitoring systems by narrowing the day-night performance gap, leveraging one-shot error correction, and reducing computational requirements.




\bibliographystyle{ieeetr}
\bibliography{weatherclassi}
\end{document}